%% file: root.tex
\documentclass[letterpaper, 10 pt, conference]{ieeeconf}

\IEEEoverridecommandlockouts
\usepackage{subcaption}
\usepackage{tabularray}
\usepackage{xcolor}
\definecolor{Gray}{gray}{0.85}
\usepackage{balance}
\usepackage{amssymb}
\usepackage[tbtags]{amsmath}
\usepackage{commath}
\usepackage{bm}
\input{h2t_def}

\usepackage{booktabs}
\usepackage{multirow}
\usepackage{graphicx}
\usepackage{fdsymbol}
\usepackage{tikz}
\usepackage{mathtools}
\usepackage{soul}
\usepackage{microtype}
\usepackage{dirtytalk}
\usepackage[noadjust]{cite}
\usepackage[hidelinks]{hyperref}
\usepackage{algorithm}
\usepackage[noend]{algpseudocode}
\usepackage{nicefrac}
\usepackage{svg}
\usepackage[abbreviations]{glossaries-extra}
\usepackage{inconsolata}
\makeatletter
\g@addto@macro\normalsize{%
  \setlength{\abovedisplayskip}{4pt plus 1pt minus 1pt}%
  \setlength{\belowdisplayskip}{4pt plus 1pt minus 1pt}%
  \setlength{\abovedisplayshortskip}{2pt plus 1pt}%
  \setlength{\belowdisplayshortskip}{2pt plus 1pt minus 1pt}%
}
\makeatother
\graphicspath{{figures/}}
\setkeys{glslink}{hyper=false}
\glssetcategoryattribute{acronym}{indexonlyfirst}{true}
\GlsXtrEnableEntryCounting{abbreviation}{1}
\makeatletter
\def\BState{\State\hskip-\ALG@thistlm}
\makeatother
\newabbreviation{lfd}{LfD}{Learning from Demonstrations}
\newabbreviation{mp}{MP}{Movement Primitive}
\newabbreviation{dmp}{DMP}{Dynamic Movement Primitive}
\newabbreviation{vmp}{VMP}{Via-Point Movement Primitive}
\newabbreviation{gmm}{GMM}{Gaussian Mixture Model}
\newabbreviation{bimacs}{Bimacs dataset}{KIT Bimanual Actions Dataset}
\newabbreviation{bmds}{BiManip dataset}{KIT Bimanual Manipulation Dataset}
\newabbreviation{bn}{BN}{Bayesian Network}
\newabbreviation{dpll}{DPLL}{Davis–Putnam–Logemann–Loveland}
\newcommand{\modintervalstart}[1]{#1^-}
\newcommand{\modintervalend}[1]{#1^+}
\newcommand{\modnext}[1]{#1'}
\newcommand{\modcandidate}[1]{\tilde{#1}}
\newcommand{\modunclear}[1]{#1^?}
\newcommand{\modinitial}[1]{#1_i}
\newcommand{\modlength}[1]{\abs{#1}}
\newcommand{\modeuclideannorm}[1]{\|#1\|_2}
\newcommand{\modeuclideannormm}[1]{\left\|#1\right\|_2}
\newcommand{\vartiming}{\tau}

\newcommand{\varscalar}{\varsigma}
\newcommand{\vartimingt}{\overset{\mathclap{\scriptscriptstyle{(3)}}}{\vartiming}}
\newcommand{\vartimingf}{\overset{\mathclap{\scriptscriptstyle{(4)}}}{\vartiming}}
\newcommand{\varDemonstrations}{\mathcal{D}}
\newcommand{\vardemonstration}{D}
\newcommand{\varActionSequence}{S}
\newcommand{\varActionSequenceLeft}{\varActionSequence_L}
\newcommand{\varActionSequenceRight}{\varActionSequence_R}
\newcommand{\varTaskAssignments}{T}
\newcommand{\varTaskAssignmentsNext}{\modnext\varTaskAssignments}
\newcommand{\varTaskAssignmentCandidates}{\modcandidate\varTaskAssignments}
\newcommand{\vartaskassignment}{t}
\newcommand{\vartaskassignmentnext}{\modnext\vartaskassignment}
\newcommand{\vartaskassignmentcandidate}{\modcandidate\vartaskassignment}
\newcommand{\vartaskassignmentinitialcandidate}{\modinitial{\modcandidate{\vartaskassignment}}}
\newcommand{\varAssigned}{B}
\newcommand{\varAssignedNext}{\modnext\varAssigned}
\newcommand{\varassigned}{b}
\newcommand{\varassignednext}{\modnext\varassigned}
\newcommand{\varUnassigned}{\modunclear\varAssigned}
\newcommand{\varUnassignedInitial}{\modinitial\varUnassigned}
\newcommand{\varunassigned}{\modunclear\varassigned}
\newcommand{\varscore}{s}
\newcommand{\varscoretaskassignment}{\varscore_t}
\newcommand{\varRelation}{R}
\newcommand{\varAllenRelSet}{\varRelation_A}
\newcommand{\varverb}{v}
\newcommand{\varobject}{o}
\newcommand{\varActions}{\mathcal{A}}
\newcommand{\varaction}{\alpha}
\newcommand{\varactiona}{\varaction_a}
\newcommand{\varactionb}{\varaction_b}
\newcommand{\varactionc}{\varaction_c}
\newcommand{\vartimedaction}{a}
\newcommand{\vartimedactionlength}{\modlength\vartimedaction}
\newcommand{\vartimedactionstart}{\modintervalstart{\vartimedaction}}
\newcommand{\vartimedactionend}{\modintervalend{\vartimedaction}}
\newcommand{\vartimedactionalength}{\modlength\vartimedactiona}
\newcommand{\vartimedactiona}{\vartimedaction_a}
\newcommand{\vartimedactionastart}{\modintervalstart{\vartimedactiona}}
\newcommand{\vartimedactionaend}{\modintervalend{\vartimedactiona}}
\newcommand{\vartimedactionb}{\vartimedaction_b}
\newcommand{\vartimedactionbstart}{\modintervalstart{\vartimedactionb}}
\newcommand{\vartimedactionbend}{\modintervalend{\vartimedactionb}}
\newcommand{\vartimedactionblength}{\modlength\vartimedactionb}
\newcommand{\vartimedactionc}{\vartimedaction_c}
\newcommand{\vartimedactioncstart}{\modintervalstart{\vartimedactionc}}

\newcommand{\vartimedactionanystart}[1]{\modintervalstart{\vartimedaction_{#1}}}
\newcommand{\vartimedactionanyend}[1]{\modintervalend{\vartimedaction_{#1}}}
\newcommand{\varinterval}{i}
\newcommand{\varintervalstart}{\modintervalstart\varinterval}
\newcommand{\varintervalend}{\modintervalend\varinterval}
\newcommand{\varthreedimspace}{\mathbb{T}^3}
\newcommand{\vartimingspacerepralength}{\lambda_{a}}
\newcommand{\vartimingspacereprblength}{\lambda_{b}}
\newcommand{\vartimingspaceoffset}{\omega_{ab}}
\newcommand{\varfourdimspace}{\mathbb{T}^4}
\newcommand{\evalexperimentalsetup}{\noindent\textbf{Experimental Setup:}\xspace}
\newcommand{\evalresults}{
\noindent\textbf{Results:}\xspace}
\newcommand{\subto}{s.\,t.\xspace}
\newcommand{\texttask}[1]{``#1''\xspace}
\newcommand{\textaction}[1]{`#1'}
\newcommand{\textrelation}[1]{\emph{#1}\xspace}
\newcommand{\textmueslitask}{\texttask{prepare muesli}}
\newcommand{\textcomponenttask}{\texttask{disassemble component}}
\newcommand{\textallenbefore}{\textrelation{before}}
\newcommand{\textallenafter}{\textrelation{after}}
\newcommand{\textallenmeets}{\textrelation{meets}}

\newcommand{\textallenoverlaps}{\textrelation{overlaps}}
\newcommand{\textallenoverlappedby}{\textrelation{overlapped by}}

\newcommand{\textallenfinishedby}{\textrelation{finished by}}
\newcommand{\textallenstarts}{\textrelation{starts}}

\newcommand{\textallenduring}{\textrelation{during}}
\newcommand{\textallenequals}{\textrelation{equals}}

\title{\LARGE \bf
Unified Learning of Temporal Task Structure and Action Timing\\for Bimanual Robot Manipulation}
\enablefinalversion
\ifx\anonymous\undefined
\author{Christian Dreher, Patrick Dormanns, Andre Meixner, and Tamim Asfour%
\thanks{The research leading to these results has received funding from the 
Deutsche Forschungsgemeinschaft (DFG, German Research Foundation) -- SFB 1574 
and the German Federal Ministry of Research, Technology and Space (BMFTR) under the Robotics Institute Germany (RIG). The authors are with the Institute for Anthropomatics and Robotics, Karlsruhe Institute of Technology, Karlsruhe, Germany. {\tt \{c.dreher,asfour\}@kit.edu}}%
}
\else
\author{Undisclosed authors for double-anonymous review%
\thanks{Acknowledgments anonymized acknowledgments anonymized acknowledgments anonymized acknowledgments anonymized acknowledgments.}
\thanks{Affiliations anonymized affiliations anonymized affiliations anonymized affiliations anonymized affiliations anonymized affiliations anonymized.}%
}
\fi

\begin{document}
\maketitle
\thispagestyle{empty}
\pagestyle{empty}

\begin{abstract}%
Bimanual manipulation requires both temporal task structure -- which actions precede or overlap others -- and concrete timing -- when each action starts and how long it takes.
Symbolic relations enable high-level reasoning about task structure and alternative execution sequences, while concrete timing parameters are essential for coordinating two hands at the execution level.
Existing approaches address these two levels in isolation, leaving a gap between task planning and movement synchronization. 
This work presents an approach for learning symbolic and subsymbolic temporal task constraints from human demonstrations and deriving executable, temporally parametrized plans for bimanual manipulation. 
Specifically, our contributions are
\begin{enumerate*}[label=(\roman*)]
    \item a 3-D representation of timings with methods based on multivariate Gaussian Mixture Models to represent temporal relationships between actions on a subsymbolic level, 
    \item a method based on the \glsentrylong{dpll} algorithm that finds and ranks all contradiction-free Allen relation assignments of a task, representing its modes, and
    \item an optimization-based planning system combining the identified symbolic and subsymbolic temporal task constraints to derive parametrized plans for robot execution.
\end{enumerate*}
We quantitatively evaluate our approach and show that the inferred symbolic task constraints are more accurate than those of prior heuristic approaches and that the generated plans are temporally closer to human demonstrations than the most characteristic demonstration.
Additionally, we qualitatively demonstrate its applicability through orchestrated task executions in simulation and on real robots.
\end{abstract}
\glsreset{dpll}

\section{Introduction}

\begin{figure}[!t]
    \vspace*{1.5mm}
    \centering
    \includegraphics[width=\linewidth]{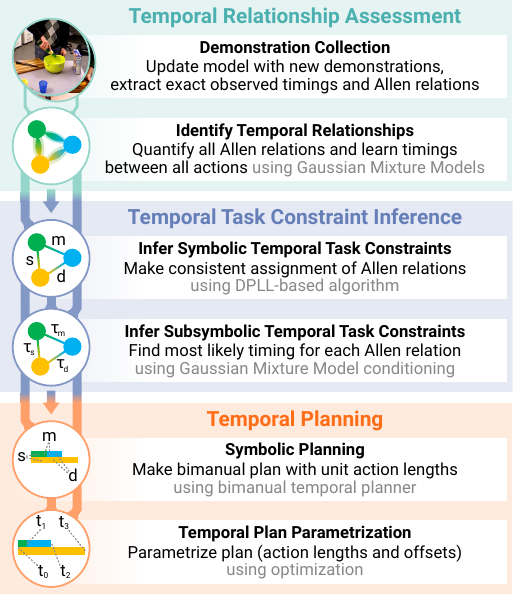}
    \vspace*{-4mm}
    \caption{Overview of our approach and contributions: \emph{Temporal Relationship Assessment} evaluates all Allen relations observed in human demonstrations, \emph{Temporal Task Constraint Inference} finds contradiction-free Allen relation assignments and most likely action timings, and \emph{Temporal Planning} generates executable, temporally parametrized bimanual plans.}
    \label{fig:graphical-abstract}
    \vspace*{-2mm}
\end{figure}
Temporal task structure and action timing are fundamental to bimanual manipulation. 
When unscrewing a component, it must be held \textallenduring unscrewing, not \textallenbefore or \textallenafter.
When pouring liquid into a container, the pouring motion should \emph{start} as the container is in position.
These temporal relationships define the logical structure of bimanual tasks. 
Yet knowing that one action occurs \textallenduring another is insufficient for execution -- we also need concrete timings: \emph{how long} does each action take, and \emph{exactly when} should each action start?
\emph{Symbolic temporal task constraints} between two actions can be modeled as qualitative relations, specifically as Allen relations~\cite{allen1983maintaining},  such as \textallenbefore, \textallenoverlaps, or \textallenduring.
They allow the robot to reason about alternative execution sequences, generate plans, and generalize to new situations. 
For instance, knowing that \textaction{transfer object} must occur \textallenafter \textaction{grasp object} but can occur \textallenbefore or \textallenduring \textaction{open container} enables flexible and context-aware scheduling.
\emph{Subsymbolic temporal task constraints} provide concrete durations, delays, and offsets that determine the execution quality -- parameters to which we refer to as \emph{timing}.
They specify whether the robot waits \SI{200}{\milli\second} for stabilization before pouring, pours for \SI{3}{\second} or \SI{5}{\second}, and how precisely the hands synchronize.

This distinction between symbolic and subsymbolic temporal information reflects two fundamentally different levels of a robot's temporal reasoning.
Yet despite their complementary roles, existing approaches largely address them in isolation.
Methods for learning task structure from demonstration focus on qualitative ordering and precedence \cite{ekvall2006learning, pardowitz2007incremental, nicolescu2003natural}, while methods for movement synchronization treat timing as a low-level control problem decoupled from task-level reasoning \cite{thota2016learning, kulvicius2013interaction, gams2014coupling, gams2015accelerating, karlsson2018convergence, krebs2024formalization}.
Our previous work \cite{dreher2024learning} addressed bridging this gap by learning both symbolic and subsymbolic temporal task constraints from demonstration.
However, it models the temporal relationship between each pair of action keypoints independently using univariate \glspl{gmm}, capturing only marginal distributions and missing the joint structure of action timings.
Additionally, it is only capable of finding one assignment of Allen relations to each pair of actions in a task and does not include a planning system to generate executable plans for the whole task.

In this work, we present a unified approach that learns both symbolic and subsymbolic temporal task constraints from human demonstrations and uses them to generate temporally parametrized plans for robot execution. 
Our approach consists of three parts, illustrated in Fig.~\ref{fig:graphical-abstract}. 
First, we introduce a 3-D embedding of timings between two actions, the \emph{timing space}, and propose multivariate \gls{gmm}-based methods that jointly model the lengths of two actions and their relative offset to capture their temporal relationship. 
Second, we propose an approach based on the \gls{dpll} algorithm~\cite{davis1962machine} that finds and ranks all contradiction-free Allen relation assignments for a task, enabling the identification of multiple task modes. 
Third, we develop an optimization-based planning system that takes the inferred symbolic and subsymbolic task constraints as input and produces a fully temporally parametrized plan that best reflects the human demonstrations.
We quantitatively evaluate our approach using the \gls{bimacs} \cite{dreher2022learning} and show that the inferred symbolic task constraints are more accurate than those of prior heuristic approaches and that the generated plans are temporally closer to human demonstrations than the most characteristic demonstration.
Additionally, we qualitatively demonstrate its applicability through orchestrated task executions in simulation and on real robots  using the \gls{bimacs} and \gls{bmds} \cite{krebs2021kit}.

\section{Related Work}

Temporal structure and timing are central to bimanual manipulation, yet the literature largely addresses the symbolic and subsymbolic levels in isolation.
Symbolic temporal task constraints are important for task planning and are discussed in Section~\ref{ss:rw-task-modes}.
Subsymbolic temporal task constraints are needed for synchronized execution of bimanual tasks and are discussed in Section~\ref{ss:rw-mp-coupling}.
Apart from the work in~\cite{dreher2024learning}, none of these works addresses the simultaneous learning of symbolic and subsymbolic temporal task constraints.

\subsection{High-level: Task Modes and Temporal Task Structure}\label{ss:rw-task-modes}

\emph{Task modes} are different sequences of key actions that lead to the same goal through varying orderings.
A \emph{key action} is any action that changes the state of the primarily manipulated objects, such as a disassembly action separating two parts.
This is referred to as \emph{temporal task structure} \cite{billard2008robot}. 
Key actions stand in contrast to supportive actions like approaching objects for grasping or lifting objects for key actions.
In the following, we discuss works on precedence relations and those that need more complex relations to handle bimanuality or simultaneous actions.

\subsubsection{Precedence Relations}\label{sss:rw-precedence}

Several early works proposed learning different task modes for unimanual tasks, or tasks with one execution sequence \cite{ekvall2006learning, pardowitz2007incremental, nicolescu2003natural}.
Carpio \etal \cite{carpio2019learning} propose a method based on \glspl{bn} to learn the sequential temporal task structure in the context of human robot interaction.
While effective for unimanual tasks, precedence-based representations do not scale naturally to bimanual tasks:  two concurrent execution sequences create a combinatorial explosion of orderings, and slight qualitative changes in supportive actions do not constitute a fundamentally different task mode.

\subsubsection{Complex Relations}\label{sss:rw-complex}

Allen \cite{allen1983maintaining} systematically defines all 13 possible relations in which two intervals can be and provides multiple deductive methods \eg to derive transitive relations.
Ye \etal~\cite{ye2019robot} consider parallel execution of independent actions using precedence graphs extended with concurrency, but this does not account for the complex temporal relationships between hands that bimanual manipulation requires.
Dreher and Asfour \cite{dreher2022learning} propose an inductive graph-based heuristic search method to learn symbolic temporal task constraints in bimanual manipulation tasks that are needed to understand task structure from partially contradictory demonstrations.
Our \gls{dpll}-based algorithm builds on and extends this work by performing an exhaustive rather than heuristic search, finding and ranking all contradiction-free Allen relation assignments rather than a single most likely one.
More recently, large language models and foundation models have been applied to high-level task planning \cite{birr2024autogptp}. 
However, these approaches operate at the level of symbolic action sequences and do not address subsymbolic timing -- they cannot specify how long an action should take or how two concurrent actions should be temporally offset.

\subsection{Low-level: MP Synchronization and Coupling}\label{ss:rw-mp-coupling}

On a lower level, bimanual coordination or synchronization can be achieved by \emph{coupling} \glspl{mp}.
This term implies that the execution of one \gls{mp} is intertwined with the execution of another \gls{mp} through some mechanism, and thus the synchronization is \emph{emerging} from it.
Another approach is to treat \glspl{mp} as black boxes and parametrize the offsets between the two \glspl{mp} of a bimanual action, as well as their duration with a targeted execution length.
In this case, the synchronization is \emph{assigned} at a higher level.

\subsubsection{Emerging Synchronization (Bottom-Up)}\label{sss:rw-tightly-coupled}

Thota \etal{} \cite{thota2016learning} propose control laws to couple \glspl{dmp} by feeding back the tracking error of one arm to the other arm in contrast to feeding it back to the arm itself to track the desired trajectory.
For this work, the formulation of \glspl{dmp} had to be adapted.
Kulvicius \etal{} \cite{kulvicius2013interaction} propose learning weights of virtual spring coupling between two end-effectors controlled by \glspl{dmp}, allowing flexible coordination to emerge from learned interaction forces.
Gams \etal{} \cite{gams2014coupling} propose to couple \glspl{dmp} in bimanual tasks with force/torque feedback learned during the robot's execution.
This was later extended to discrete-periodic motion on a humanoid robot, where phase coupling between primitives drives synchronization~\cite{gams2015accelerating}.
Karlsson \etal{} \cite{karlsson2018convergence} introduce temporal coupling between \glspl{dmp} through a shared phase variable, ensuring convergence of both primitives regardless of execution speed variations.
Although these bottom-up approaches yield robust synchronization, they require task-specific coupling designs and do not generalize the temporal structure of a task across different execution modes or new situations.
A different line of work bypasses explicit coupling altogether by learning end-to-end bimanual policies from demonstrations, \eg transformer-based action-chunking~\cite{motoda2025learning}.
Such approaches are reactive, but do not expose interpretable task constraints and require many demonstrations.

\subsubsection{Assigned Synchronization (Top-Down)}\label{sss:rw-loosely-coupled}

An alternative is to treat MPs as black boxes and assign synchronization parameters (offsets and durations) from a higher level.
The works in  \cite{zoellner2004programming} and \cite{krebs2024formalization} achieve this in the context of bimanual manipulation using pre-defined Petri nets that encode the temporal structure of bimanual manipulation categories.
These approaches are expressive, but require the temporal structure to be hand-designed rather than learned.
The work in~\cite{dreher2024learning} is closest to our work. 
The authors propose a method for extracting symbolic and subsymbolic temporal task constraints from human demonstrations, where univariate \glspl{gmm} are used to learn temporal differences between semantic action keypoints (starts and ends of actions).
This approach is data-efficient but considers each pair of keypoints independently, missing the joint structure of action timings. 
In particular, it cannot capture correlations between action lengths and offsets, which are critical for generating consistent parametrized plans.
In this work, we propose a multivariate \gls{gmm} formulation to fully capture the temporal relationship between two actions.

To our knowledge, no existing approach simultaneously 
\begin{enumerate*}[label=(\roman*)]
    \item learns symbolic temporal task structure at the level of Allen relations from demonstrations that may exhibit multiple task modes, 
    \item models the full joint distribution of action timings using a principled representation, and 
    \item integrates both levels into a single planning system that produces temporally parametrized plans for execution.
\end{enumerate*}

\section{Problem Statement}

In this section, we first introduce basic formalisms and definitions required in the scope of this work.

\subsection{Definition of Task, Action, Demonstration}\label{ss:def-task-action-demonstration}

The following definitions are largely based on our previous work \cite{dreher2024learning} with minor modifications and additions. 
A bimanual manipulation \emph{task} consists of a set of actions $\varActions$, with $\varActions \! = \! \{\varactiona, \! \varactionb, \! \ldots\}$ and where an \emph{action} $\varaction$ is a tuple $\varaction \! = \! (\varverb, \! \varobject)$, with $\varverb$ denoting a verb and $\varobject$ an object.
For actions enriched with temporal information -- relevant when observing or reproducing an action with concrete start and end -- the notation $\vartimedaction\!=\!(\varverb, \varobject, \varinterval)$ is used, where $\varinterval \! = \! [\varintervalstart \! , \! \varintervalend]$ is a time interval with start $\varintervalstart$ and end $\varintervalend \!>\!\varintervalstart$. 
We write $\vartimedactionstart \!\! = \! \varintervalstart$ and $\vartimedactionend \!\! = \! \varintervalend$ to refer to the two semantic temporal keypoints of the action and define its \emph{length} as $\vartimedactionlength \! = \! \vartimedactionend \! - \! \vartimedactionstart$.
The \emph{offset} $ \vartimingspaceoffset $ between two actions $\vartimedactiona$ and $\vartimedactionb$ is defined as the distance between their midpoints: $ \vartimingspaceoffset\!=\!\frac{1}{2}(\vartimedactionbstart \! + \! \vartimedactionbend) \! - \! \frac{1}{2}(\vartimedactionastart \! + \! \vartimedactionaend) $.
Let $\varDemonstrations$ be a set of demonstrations of a bimanual manipulation task. A \emph{demonstration} $\vardemonstration \! \in \! \varDemonstrations$ is a tuple $\vardemonstration\!=\!(\varActionSequenceLeft, \! \varActionSequenceRight)$ consisting of two action sequences $\varActionSequenceLeft$ for the left hand, and $\varActionSequenceRight$ for the right hand, respectively.
An \emph{action sequence} $\varActionSequence$ is an ordered collection of actions $\varActionSequence \! = \! (\vartimedactiona, \! \vartimedactionb, \! \vartimedactionc, \! \ldots)$.
Actions $\vartimedactiona, \! \vartimedactionb, \! \vartimedactionc, \! \ldots$ in an action sequence $\varActionSequence \! = \!(\vartimedactiona, \! \vartimedactionb, \! \vartimedactionc, \! \ldots)$ have a strict order and cannot overlap, thus it follows that $\vartimedactionastart\!<\!\vartimedactionaend\!\leq\!\vartimedactionbstart\!<\!\vartimedactionbend\!\leq\!\vartimedactioncstart \ldots$ etc.
A task consists of several subtasks, \ie subsets of actions showing high temporal cohesion (symbolically or subsymbolically) \cite{dreher2022learning}.
A set of demonstrations $\varDemonstrations$ may exhibit multiple \emph{task modes}, which are qualitatively different but equally valid orderings of subtasks.
Additionally, each subtask may show different modes that manifest as slight variations of concurrent temporal relations, but rarely different sequences.
Both types of modes reflect the natural variation in how humans perform the same task.

\subsection{Definition of Task Assignments}

A single set of demonstrations may contain various modes.
This results in apparent contradictions when temporal relations between action pairs are unconditionally assessed. 
A robot needs to find one arrangement of all the actions in the task on a symbolic level so that they are consistent as a whole, and most likely given the demonstrations.
To this end, we define a \emph{task assignment} $\vartaskassignment \! = \! (\varAssigned, \! \varscoretaskassignment)$, where $\varAssigned$ is a contradiction-free complete assignment of one single Allen relation $\varRelation \in \varAllenRelSet$ to each action pair $(\varactiona, \! \varactionb) \! \in \! \varActions \! \times \! \varActions \! : \! \varactiona \! \neq \! \varactionb$, $\varAllenRelSet$ is the set of all $13$ Allen relations~\cite{allen1983maintaining}, and a score $\varscoretaskassignment$ quantifying how well the assignment reflects the observed demonstrations. 
Formally, $\varAssigned$ is a set of 4-tuples $\varassigned \! = \! (\varactiona, \! \varactionb, \! \varRelation, \! \varscore)$, with $(\varactiona, \! \varactionb) \! \in \! \varActions \! \times \! \varActions \! : \! \varactiona \! \neq \! \varactionb$ being an action pair where $\varRelation$ is the Allen relation assigned to the action pair $(\varactiona, \! \varactionb)$, and $\varscore \! \in  \! [0, \! 1]$ being a value that expresses the degree to which the relation $\varactiona \varRelation \varactionb$ is supported by the demonstrations.
We also consider a \emph{partial task assignment} $\vartaskassignmentcandidate$ as a tuple $\vartaskassignmentcandidate \! = \! (\varUnassigned, \varAssigned)$ as a means to obtain a complete task assignment $\vartaskassignment$ in the proposed algorithms.
Here, $\varAssigned$ is the set of concrete assignments of Allen relations to action pairs 
and $\varUnassigned$ is the set of all action pairs that have not yet been assigned.
An unassigned action pair $\varunassigned \! \in \! \varUnassigned$ is a tuple $\varunassigned \! = \! (\varactiona, \! \varactionb)$ with $(\varactiona, \! \varactionb) \! \in \! \varActions \! \times \! \varActions \! : \! \varactiona \! \neq \! \varactionb$.
A task assignment makes a temporal relation a symbolic temporal task constraint because it guarantees that a single relation will not cause contradictions.
In this work, our first aim is to identify all contradiction-free task assignments and rank them by their score $\varscoretaskassignment$, \ie likeliness given the demonstrations.

\subsection{Definition of Timing}

A task assignment captures the qualitative temporal structure of a task, but is insufficient for robot execution, which requires concrete timing parameters. 
We define the \emph{timing} $\vartiming$ between two actions $\vartimedactiona$ and $\vartimedactionb$ as a compact representation of their concrete temporal relationship.
Throughout this work, two equivalent representations will be used.
Naively, the timing between the two actions can be represented as 4-D vector $\vartimingf \! = \! (\vartimedactionastart, \! \vartimedactionaend, \! \vartimedactionbstart, \! \vartimedactionbend), \vartimingf \in \varfourdimspace$, with the components being the starts and ends of both actions.
However, a 3-D vector representation $\vartimingt \! = \! (\vartimingspacerepralength, \! \vartimingspacereprblength, \! \vartimingspaceoffset), \vartimingt \in \varthreedimspace$ can also be used, where $\vartimingspacerepralength$ and $\vartimingspacereprblength$ correspond to the lengths of actions $\vartimedactiona$ and $\vartimedactionb$ respectively, and $\vartimingspaceoffset$ to the offset between both actions.
In the context of this work, these representations $\vartimingf$ and $\vartimingt$ can be used interchangeably, as they can be converted into each other.
The $\vartimingt$ form can be preferable for learning since it is invariant to uniform time shifts.
If the concrete representation is not important, we use the notation $\vartiming$ for brevity.
To learn from several demonstrations, we need a model that represents the timing between two actions.
Thus, we define a \emph{timing model} as a multivariate \gls{gmm} in timing space.
A subsymbolic temporal task constraint is the most likely timing between two actions given a specific symbolic temporal task constraint, \ie given the assigned Allen relation.
In this work, our second aim is to infer subsymbolic temporal task constraints from timing models.

\subsection{Definition of Plan and Temporally Parametrized Plan}

A \emph{plan} in the context of this work is a pair of action sequences $ (\varActionSequenceLeft, \varActionSequenceRight)$ for the left and right hand, in which actions are qualitatively arranged to satisfy symbolic temporal task constraints.
Since it is the output of a symbolic planner, it exhibits discrete and artificial timings.
A \emph{temporally parametrized plan} extends this by assigning concrete lengths and offsets to all actions, consistent with the symbolic constraints and as close as possible to the inferred subsymbolic temporal task constraints.
Both are denoted by $P \! = \! (\varActionSequenceLeft, \! \varActionSequenceRight)$, are equivalent to a demonstration, and differ only in integer or real timing values.
In this work, our third aim is to derive a temporally parametrized plan given identified symbolic and subsymbolic temporal task constraints for robot execution.

\section{Approach}

This section describes our approach with its three parts also illustrated in Fig.~\ref{fig:graphical-abstract}.

\subsection{Temporal Relationship Assessment}\label{ss:temporal-relationship-assessment}

Assessing temporal relationships between actions symbolically and subsymbolically is the cornerstone of our approach.

\subsubsection{Assessing Temporal Relations}\label{sss:assess-temp-rel}

Temporal relations provide a qualitative source of temporal information needed in our approach.
To assess which temporal relations between two actions are most likely to hold given all observed demonstrations, we adopt the method described in \cite{dreher2024learning}, which quantifies the degree to which each Allen relation holds between each action pair using univariate \glspl{gmm} and fuzzy logic. 
The result is a soft, graded assessment of temporal relations for every action pair in the task. 

\subsubsection{Assessing Timing}\label{sss:assess-timing}

\begin{figure*}[t!]
    \centering
    \begin{minipage}[t]{0.32\textwidth}
        \centering
        \begin{minipage}[c][0.62\linewidth][c]{\linewidth}
            \centering
            \def\svgwidth{0.85\linewidth}
            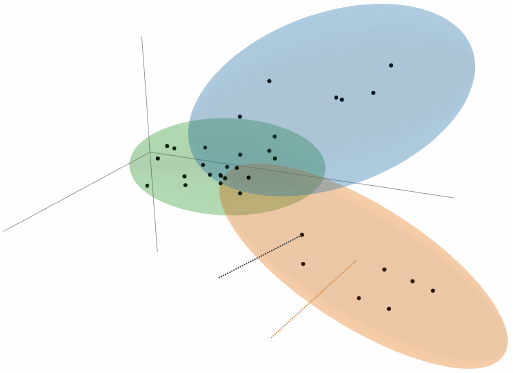
        \end{minipage}
        \vspace*{-1mm}
        \caption{
            Timing model in $\varthreedimspace$, modeling the relationship between two actions.
            Black dots are observed timings from several human demonstrations.
            Ellipsoids denote modes of the multivariate \gls{gmm}.%
        }\label{fig:gmms}
    \end{minipage}
    \hfill
    \begin{minipage}[t]{0.32\textwidth}
        \centering
        \begin{minipage}[c][0.62\linewidth][c]{\linewidth}
            \centering
            \def\svgwidth{0.85\linewidth}
            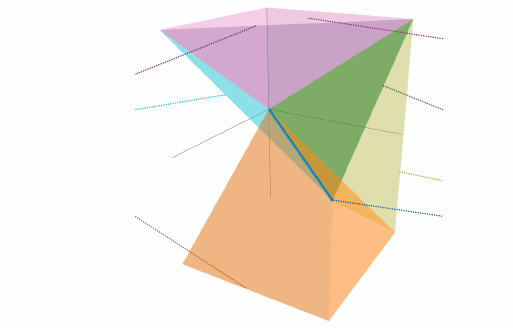
        \end{minipage}
        \vspace*{-1mm}
        \caption{%
            \vphantom{$\varthreedimspace$}Visualization of selected Allen relations in $\varthreedimspace$ as line (\textallenequals), areas (\eg \textallenfinishedby) or volumes (\eg \textallenbefore).
            Points in these regions can be attributed to the respective Allen relation.%
        }\label{fig:allen_relations_in_t3}
    \end{minipage}
    \hfill
    \begin{minipage}[t]{0.32\textwidth}
        \centering
        \begin{minipage}[c][0.62\linewidth][c]{\linewidth}
            \centering
            \def\svgwidth{0.85\linewidth}
            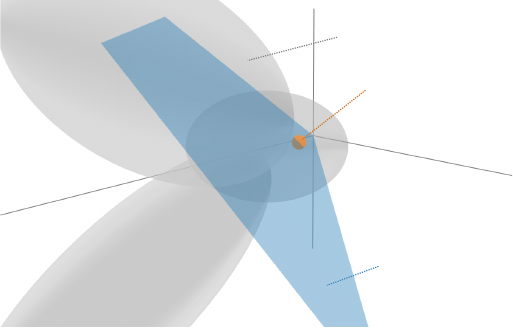
        \end{minipage}
        \vspace*{-1mm}
        \caption{%
            \vphantom{$\varthreedimspace$}Conditioning a timing model to the Allen relation \textallenfinishedby (blue area).
            The orange point is the maximum of the \gls{gmm}'s probability density function restricted to that area.%
        }\label{fig:gmm-conditioning}
    \end{minipage}
    \vspace*{-5mm}
\end{figure*}

Beyond qualitative relations, execution requires concrete timing information.   
We represent the timing between two actions $\vartimedactiona$ and $\vartimedactionb$ of a task as a 3-D vector $(\vartimingspacerepralength, \vartimingspacereprblength, \vartimingspaceoffset)$, where $\vartimingspacerepralength$ and $\vartimingspacereprblength$ correspond to the lengths of the actions and $\vartimingspaceoffset$ to their midpoints' offset.
All observed timings between two actions are expressed as such vectors and embedded in timing space $\varthreedimspace$ to train multivariate Bayesian \glspl{gmm}, whose variational formulation automatically determines the required number of mixture components.
Fig.~\ref{fig:gmms} shows a visualization of the timing space with several demonstrations of \textaction{pour milk} and \textaction{hold bowl} together with the corresponding multivariate \gls{gmm} that models the timing distribution.
Allen relations can conveniently be represented in timing space by lines, areas, or volumes, as shown in Fig.~\ref{fig:allen_relations_in_t3}.

The 3-D space $\varthreedimspace$ is better suited for representing observations, since it reduces the problem space for learning by ignoring shifts of both intervals by the same amount as is the case in $\varfourdimspace$, \ie shifts along $(1, \! 1, \! 1, \! 1)$.
Consider the two timings $(1, \! 2, \! 3, \! 4) \! \neq \! (2, \! 3, \! 4, \! 5)$: In $\varfourdimspace$, they are two different points even though they describe the exact same relative temporal arrangement between actions.
In $\varthreedimspace$, both will be represented as the same point, accurately reflecting only their relationship with each other.
This is also a benefit when considering norms and distances in $\varthreedimspace$, which is important for our approach.
However, by naively using the lengths of two actions, \ie $(\vartimingspacerepralength, \! \vartimingspacereprblength, \! \vartimingspaceoffset) \! = \! (\vartimedactionalength \!\! , \vartimedactionblength \!\! , \! \vartimingspaceoffset)$, we would render the Euclidean norm $\modeuclideannorm{\vartimingt}$ for $\vartimingt \! \in \! \varthreedimspace$ meaningless.
Instead, since we are only interested in relative temporal arrangements as just discussed, we want the Euclidean norm $\modeuclideannorm{\vartimingt}$ to be equal to the minimum Euclidean norm $\modeuclideannorm{\vartimingf}$ considering shifts by the same amount:
\begin{equation}%
    \modeuclideannorm{\vartimingt}\!=\!
    \min_\varscalar \modeuclideannormm{\vartimingf + \varscalar \begin{pmatrix} 1 \\ 1 \\ 1 \\ 1 \end{pmatrix}}\!
    , \text{with } 
    \vartimingt\!=\!\begin{pmatrix} \vartimingspacerepralength \\ \vartimingspacereprblength \\ \vartimingspaceoffset \end{pmatrix}\!
    ,
    \vartimingf\!=\!\begin{pmatrix} \vartimedactionastart \\ \vartimedactionaend \\ \vartimedactionbstart \\ \vartimedactionbend \end{pmatrix}\!
    .
\end{equation}%
More precisely, with a scalar $s$ and $\vartimingspacerepralength=s\vartimedactionalength$ and $\vartimingspacereprblength=s\vartimedactionblength$, we require that %
\begin{equation}%
    \modeuclideannormm{\begin{pmatrix} s\vartimedactionalength \\ s\vartimedactionblength \\ \vartimingspaceoffset \end{pmatrix}}\!=\!
    \min_\varscalar \modeuclideannormm{\begin{pmatrix} \vartimedactionastart \\ \vartimedactionaend \\ \vartimedactionbstart \\ \vartimedactionbend \end{pmatrix} + \varscalar \begin{pmatrix} 1 \\ 1 \\ 1 \\ 1 \end{pmatrix}}
    .
\end{equation}%
Solving for $s$ yields $s=\frac{1}{\sqrt{2}}$. It follows that we embed a timing into our 3-D space $\varthreedimspace$ as follows:
\begin{equation}%
    \vartimingt = \begin{pmatrix}\frac{1}{\sqrt{2}} \vartimedactionalength\!,
                                 \frac{1}{\sqrt{2}} \vartimedactionblength\!,
                                 \vartimingspaceoffset \end{pmatrix}^{\mathsf{T}}
    .
    \label{eq:three-dim-repr}
\end{equation}%
Thus, using this embedding, we obtain a meaningful Euclidean norm in $\varthreedimspace$ that replicates the Euclidean norm in $\varfourdimspace$ if shifts along $(1, \! 1, \! 1, \! 1)$ are ignored.
The distance $d({\vartiming_1, \vartiming_2})$ between two timings $\vartiming_1$ and $\vartiming_2$ is then defined as
\begin{equation}
	d({\vartiming_1, \vartiming_2})\!=%
    \!\|\vartimingt_2\!-\!\vartimingt_1\|_2\!=%
    \!\min_\varscalar\!\left\| \begin{pmatrix}\!\vartimedactionanystart{a_2}\!\\\!\vartimedactionanyend{a_2}\!\\\!\vartimedactionanystart{b_2}\!\\\!\vartimedactionanyend{b_2}\!\end{pmatrix} \! 
    - \! %
    \left( \!\! %
    \begin{pmatrix}\!\vartimedactionanystart{a_1}\!\\\!\vartimedactionanyend{a_1}\!\\\!\vartimedactionanystart{b_1}\!\\\!\vartimedactionanyend{b_1}\!\end{pmatrix} \! %
    + \! 
    \varscalar \! 
    \begin{pmatrix} 1 \\ 1 \\ 1 \\ 1 \end{pmatrix} \!\! %
    \right) %
    \right\|_2
    .
    \label{eq:timing-distance}
\end{equation}
To conclude, our proposed 3-D timing space $\varthreedimspace$ is isomorphic to the quotient space $\varfourdimspace/ \langle (1, 1, 1, 1) \rangle$.
In this space, we train \glspl{gmm} which form the basis of the subsymbolic temporal task constraints that we obtain later.

\subsection{Temporal Task Constraint Inference}\label{ss:temporal-task-constraint-inference}

After assessing the relationships between actions, we need to infer symbolic and subsymbolic temporal task constraints that can be used for planning task executions on a robot.

\subsubsection{Infer Symbolic Temporal Task Constraints}\label{sss:infer-symbolic-temporal-task-constraints}

In the previous part (Section~\ref{sss:assess-temp-rel}), temporal relations have been assessed on an action pair basis using our previous approach~\cite{dreher2024learning}.
However, this information is a superposition of all demonstrations and is not very useful yet for the execution on a robot, as it contains many contradictions from differently observed timings or even different task modes.
Thus, we propose a \gls{dpll}-based approach that is able to find all contradiction-free assignments of Allen relations to the task and rank them by likeliness.
This is a hard problem, since it is a $13$-ary satisfiability problem, thus being NP-complete.
Algorithm~\ref{alg:find-assignments} shows the general outline.
An initial task assignment $\vartaskassignmentinitialcandidate$ is created with $\vartaskassignmentinitialcandidate \! = \! (\varUnassignedInitial, \! \emptyset)$ on line \ref{algln:find-assignments-initial-partial-task-assignment}.
Here, $\varUnassignedInitial$ is the exhaustive set of all action pairs in the task, \ie the Cartesian product of the set of all actions in the task without identical pairs.
With $\vartaskassignmentinitialcandidate$ as the starting point, the set of all partial task assignments is initialized on line \ref{algln:find-assignments-partial-task-assignments}.
The main loop on line \ref{algln:find-assignments-main-loop} continuously pops a partial task assignment $\vartaskassignmentcandidate$ from $\varTaskAssignmentCandidates$, from which next possible assignments will be derived if the task assignment is not complete yet.
A task assignment is not complete if there is at least one action pair to which no Allen relation has been assigned yet (line \ref{algln:find-assignments-if-not-complete}).
Specifically, on line \ref{algln:find-assignments-assign-next}, one specific action pair will be chosen and all feasible Allen relations will be assigned to it using Algorithm \ref{alg:assign-next}.
This creates new partial task assignments for each possible Allen relation for the chosen action pair, which are then pushed to $\varTaskAssignmentCandidates$.
\begin{algorithm}
\footnotesize
\caption{Find all feasible contradiction-free (contr.-free) task assignments sorted by highest score first.}\label{alg:find-assignments}
\begin{algorithmic}[1]
\Function{FindAssignments}{}
    \State $\varUnassignedInitial \gets \{ (\varactiona, \varactionb) \in \varActions \times \varActions \mid \varactiona \neq \varactionb \}$
    \State $\vartaskassignmentinitialcandidate \gets (\varUnassignedInitial, \emptyset)$ \label{algln:find-assignments-initial-partial-task-assignment}  
        \Comment initial partial task assignment
    \State $\varTaskAssignmentCandidates \gets \{\vartaskassignmentinitialcandidate\}$ \label{algln:find-assignments-partial-task-assignments} 
        \Comment partial task assignments (stack)
    \State $\varTaskAssignments \gets \emptyset$ \Comment contr.-free complete task assignments
    \While {$\varTaskAssignmentCandidates \neq \emptyset$} \label{algln:find-assignments-main-loop} 
        \State pop $\vartaskassignmentcandidate = (\varUnassigned, \varAssigned)$ from $\varTaskAssignmentCandidates$ 
        \If {$\varUnassigned \neq \emptyset$} \label{algln:find-assignments-if-not-complete} \Comment{if any unassigned}
            \State $\varTaskAssignmentCandidates \gets \Call{AssignNext}{\vartaskassignmentcandidate} \cup \varTaskAssignmentCandidates$ \label{algln:find-assignments-assign-next}
        \Else \Comment{if all assigned}
            \State $\varscoretaskassignment \gets \Call{ScoreAssignment}{\varAssigned}$ \label{algln:find-assignments-scoring} \Comment score
            \State push $\vartaskassignment = (\varAssigned, \varscoretaskassignment)$ to $\varTaskAssignments$ \label{algln:find-assignments-push-complete}
        \EndIf
    \EndWhile
    \State \Call{Sort}{$\varTaskAssignments$} \label{algln:find-assignments-sorted} \Comment sort descending with score
    \State \Return $\varTaskAssignments$ \label{algln:find-assignments-return}
\EndFunction
\end{algorithmic}
\end{algorithm}
If a popped task assignment turns out to be complete on line \ref{algln:find-assignments-if-not-complete}, it will be scored (line \ref{algln:find-assignments-scoring}, using Algorithm \ref{alg:score-assignment}) and pushed to the set of complete task assignments $\varTaskAssignments$ (line \ref{algln:find-assignments-push-complete}).
The sorted set (line \ref{algln:find-assignments-sorted}) of complete and contradiction-free task assignments will be returned (line \ref{algln:find-assignments-return}) after $\varTaskAssignmentCandidates$ no longer contains any task assignment with any action pair that has not been assigned to an Allen relation.

Algorithm~\ref{alg:assign-next} shows how several new partial or complete task assignments are created, given a partial task assignment, by choosing one action pair and trying to assign each Allen relation to it.
Each feasible assignment (identified on line \ref{algln:assign-next-is-feasible} using Algorithm \ref{alg:is-feasible}) of an Allen relation to an action pair will result in a new task assignment returned by this function.
Note that this function may return an empty set if no Allen relation can be assigned to the selected action pair without contradictions, causing Algorithm \ref{alg:find-assignments} to backtrack.
\begin{algorithm}[h]
\caption{Create next contradiction-free partial or complete task assignments from the given current partial task assignment $\vartaskassignmentcandidate$ by picking one unassigned action pair and assigning each feasible Allen relation.}\label{alg:assign-next}
\begin{algorithmic}[1]
\footnotesize
\Function{AssignNext}{$\vartaskassignmentcandidate = (\varUnassigned, \varAssigned)$}
    \State $\varTaskAssignmentsNext \gets \emptyset$ \Comment next task assignments
    \State pop $\varunassigned = (\varactiona, \varactionb)$ from $\varUnassigned$
    \ForAll {$\varRelation \in \varAllenRelSet$} \Comment for each Allen relation
        \State $\varscore \gets \Call{Score}{\varactiona, \varactionb, \varRelation} $ \Comment score for $\varactiona~\varRelation~\varactionb$
        \State $\varassignednext \gets (\varactiona, \varactionb, \varRelation, \varscore)$
        \If {$\Call{IsFeasible}{\varAssigned, \varassignednext}$} \label{algln:assign-next-is-feasible}
            \State $\varAssignedNext \gets \varAssigned \cup \{\varassignednext\}$ \Comment next assigned
            \State push $\vartaskassignmentnext = (\varUnassigned, \varAssignedNext)$ to $\varTaskAssignmentsNext$
        \EndIf
    \EndFor
    \State \Return $\varTaskAssignmentsNext$
\EndFunction
\end{algorithmic}
\end{algorithm}

Algorithm \ref{alg:is-feasible} outlines the procedure to ensure that the given partial assignment $\varAssigned$ would still be feasible and free of contradictions if the next assignment $\varassignednext$ is added to $\varAssigned$.
The next assignment $\varassignednext$ is a 4-tuple with $\varassignednext \! = \! (\varactiona, \! \varactionc, \! \varRelation, \! \varscore)$, \ie an assignment of the Allen relation $\varRelation$ with a score $\varscore$ between action $\varactiona$ and $\varactionc$.
In this work, we use the degree of membership of an Allen relation between two actions given the demonstrations \cite{dreher2024learning}.
Let $\varRelation_1 \! \circ \! \varRelation_2 \! \subseteq \! \varAllenRelSet$ be the composition of Allen relations, \ie the set of relations that may hold between $\varactiona$ and $\varactionc$ given $\varactiona \varRelation_1 \varactionb$ and $\varactionb \varRelation_2 \varactionc$, as defined by the transitivity table of Allen~\cite[Fig.~4]{allen1983maintaining}.
\begin{algorithm}[b]
\footnotesize
\caption{Ensure that the given partial assignment $\varAssigned$ would still be consistent with the next assignment $\varassignednext$ according to the transitivity table of Allen \cite[Fig.~4]{allen1983maintaining}.}\label{alg:is-feasible}
\begin{algorithmic}[1]
\Function{IsFeasible}{$\varAssigned, \varassignednext = (\varactiona, \varactionc, \varRelation, \varscore)$}
    \State $I \gets \{ (\varactionb, \varRelation_1, \varRelation_2) \mid \exists \varactionb \in \varActions: $
    \State \hspace{\algorithmicindent}\hspace{\algorithmicindent}$(\varactiona, \varactionb, \varRelation_{1}, \varscore_1) \in \varAssigned, (\varactionb, \varactionc, \varRelation_{2}, \varscore_2) \in \varAssigned \}$
    \ForAll {$(\varactionb, \varRelation_1, \varRelation_2) \in I$}
        \If {$\varRelation \notin \varRelation_1 \circ \varRelation_2$} \Comment{with $\varactiona \varRelation \varactionc$, $\varactiona \varRelation_1 \varactionb$, $\varactionb \varRelation_2 \varactionc$}
            \State \Return \textbf{false}
        \EndIf
    \EndFor
    \State \Return \textbf{true}
\EndFunction
\end{algorithmic}
\end{algorithm}
Then, to evaluate whether $\varassignednext$ can be added without causing contradictions in $\varAssigned$, the algorithm verifies for each action $\varactionb \! \in \! \varActions \! : \! \varactiona \varRelation_1 \varactionb \! \wedge \! \varactionb \varRelation_2 \varactionc$ that $\varRelation \! \in \! \varRelation_1 \! \circ \! \varRelation_2$ holds, \ie that $\varactiona \varRelation \varactionc$ is consistent with all already established relations involving a shared action $\varactionb$.

Algorithm \ref{alg:score-assignment} shows the scoring function for a complete task assignment.
It evaluates the sum of all individual assignment scores $\varscore$ with $\varassignednext \! = \! (\varactiona, \! \varactionc, \! \varRelation, \! \varscore)$ whose corresponding assigned relation $\varRelation$ is an overlapping Allen relation, \ie any Allen relation except for \textallenbefore or \textallenafter.
We exclude them from scoring because the assignment is solved per subtask, where precedence constraints do not contribute and concurrent relations carry the important information on action cohesion.
\begin{algorithm}[h]
\footnotesize
\caption{Score a complete task assignment $\varAssigned$.}\label{alg:score-assignment}
\begin{algorithmic}[1]
\Function{ScoreAssignment}{$\varAssigned$}
    \State $S \gets \{ \varscore \! \mid \! (\varactiona, \varactionb, \varRelation, \varscore) \in \varAssigned \wedge \varRelation$ is an overlapping Allen relation $\}$
    \State \Return $\sum{S}$
\EndFunction
\end{algorithmic}
\end{algorithm}

The heuristic rule-based approach from our previous work~\cite{dreher2022learning} is used to decompose the task into subtasks.
Our proposed \gls{dpll}-based algorithm combined with this heuristic enables us to identify task modes in the sense of variations in a subtask, and task modes in the sense of macroscopic variations of subtask sequences in a task.
As in~\cite{dreher2022learning}, we assume that subtasks are constrained by precedence constraints only, or not at all.
We argue this assumption is weak, as it merely leverages the demonstrator's bimanuality: a human cannot perform two subtasks requiring tight temporal coordination at once, and actions that do require such coordination form a single subtask by construction.

The proposed algorithm has a time complexity of $ O(13^n) $, with $13$ Allen relations that can be assigned to each action pair, $n$ being the total number of action pairs with $ n = \frac{\abs{\varActions}(\abs{\varActions}-1)}{2} $, and $\abs{\varActions}$ being the number of actions in that subtask.
Apart from the divide-and-conquer strategy of decomposing the task into subtasks, another measure based on~\cite{dreher2022learning} can be taken to make the proposed algorithm tractable in practice.
The relations \textallenmeets and \textallenbefore (and their inverses) can often already be pre-assigned due to unambiguously high scores for a specific relation, effectively reducing the dimensionality of the search space by one for each pre-assigned relation.

\subsubsection{Infer Subsymbolic Temporal Task Constraints}\label{sss:infer-subsymbolic-temporal-task-constraints}

With our approach proposed in Section~\ref{sss:infer-symbolic-temporal-task-constraints}, we obtain a temporal task assignment, \ie one concrete Allen relation per action pair in the task.
A timing model as proposed in Section~\ref{sss:assess-timing} contains the collective timing knowledge of an action pair for all Allen relations, but we can retrieve an optimal timing for a given Allen relation by conditioning.
To this end, for Allen relations that impose any time point equality (\ie lines or areas in timing space), we sample the \gls{gmm}'s probability density, orthogonally project the samples to the respective Allen relation's region in $\varthreedimspace$ and select the sample of maximum likelihood.
In the case of an Allen relation without time point equalities (\ie volumes in timing space), we just directly sample from the \gls{gmm} in that region.
An example of this process is illustrated in Fig.~\ref{fig:gmm-conditioning}.

\subsection{Temporal Planning}\label{ss:temporal-planning}

After the inference process, symbolic and subsymbolic temporal task constraints are organized in a graph structure.
However, a graph cannot directly be used for the task execution on a robot, hence a plan must be drawn up that follows the symbolic temporal task constraints tightly (Section~\ref{sss:symbolic-planning}) and the subsymbolic temporal task constraints as well as possible (Section~\ref{sss:temporal-plan-parametrization}).

\begin{figure*}[t!]
    \centering
    \begin{subfigure}[t!]{0.596\linewidth}
        \centering
        \includegraphics[width=\textwidth]{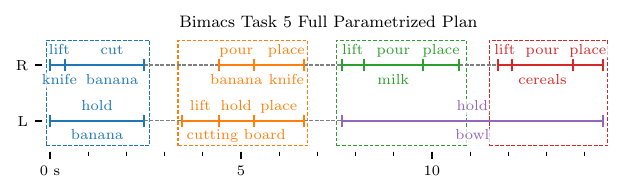}
    \end{subfigure}%
    \hfill
    \begin{subfigure}[t!]{0.396\linewidth}
        \centering
        \includegraphics[width=\textwidth]{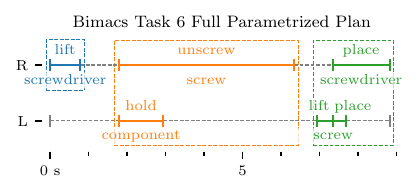}
    \end{subfigure}
    \vspace*{-3mm}
    \caption{Full temporally parametrized plans of two complex tasks of the \glsxtrshort{bimacs} generated by our approach. Task 5, \textmueslitask on the left, and Task 6, \textcomponenttask on the right.
    Boxes depict identified subtasks, with \textaction{hold bowl} being an exception, showing that actions can span subtasks.}\label{fig:fully-parametrized-tasks}
    \vspace*{-6mm}
\end{figure*}

\subsubsection{Symbolic Planning}\label{sss:symbolic-planning}

In~\cite{dreher2022learning} we developed a bimanual temporal planner to generate a synthetic dataset of bimanual manipulation demonstrations of a task by varying the sequences of actions with predefined lengths until a set of symbolic temporal task constraints is met.
We extended this planner to additionally vary the lengths of actions in multiples of unit-lengths until qualitative solutions are found that satisfy the symbolic temporal task constraints at hand.

\subsubsection{Temporal Plan Parametrization}\label{sss:temporal-plan-parametrization}

The symbolic plan will be parametrized to meet the inferred subsymbolic temporal task constraints as well as possible.
For this, we formulate an optimization problem where the symbolic temporal plan serves as hard constraints and the subsymbolic temporal task constraints as soft constraints.
Figuratively, the optimizer takes the symbolic plan as a blueprint and shifts the action's starts and ends so that the timings in the plan are as close to the inferred subsymbolic temporal task constraints as possible.
Specifically, let $\vartiming_i$ be the timing and $R_{\vartiming_i}$ the respective Allen relation of the $i$-th action pair inferred from our approach, and let $x_i$ be the corresponding timing of the same actions in the plan to optimize, with $i \! \in \! \{1, \! \dots, \! n\}$, where $n$ is the total number of action pairs considered.
Additionally, let $R_{x_i}$ be the Allen relation the timing $x_i$ resembles.
Then the objective is to minimize the Euclidean norm of the vector of all timing differences, \subto $x \! \in \! \Omega$, with decision variable $x \! = \! (x_1, \! x_2, \! \dots, \! x_n)$ and constraint set $\Omega \! = \! \{ x \! \mid \! \forall i \in \{1, \dots, n\} \! : \! R_{x_i} \! = \! R_{\vartiming_i} \}$ as follows:
\newcommand{\optterm}[1]{d(\vartiming_{#1}, x_{#1})}
\begin{equation}
    \min_{x} \modeuclideannormm{ \begin{pmatrix}\optterm{1}, \dots, \optterm{n} \end{pmatrix}^{\mathsf{T}} }
    \text{\subto~} x \in \Omega %
    .
\end{equation}
We use a standard convex optimization problem solver (CVXPY) to find an optimal solution.
The result is a fully temporally parametrized plan that can be used to parametrize the concrete durations and offsets of \glspl{mp} for robot execution.

\section{Experiments and Evaluation}

In this section, we will describe our evaluation and experiments and report the results.

\subsection{Task Assignment Assessment}

\evalexperimentalsetup To quantitatively evaluate our \gls{dpll}-based approach (``Fuzzy + \gls{dpll}-based''), we compare its inferred symbolic temporal task constraints to those of the rule-based heuristic from~\cite{dreher2022learning} in two variants: using relative frequencies of observed Allen relations as input (``Rel. Freq. + Heuristic''), and using the fuzzy-logic-based formulation of Allen relations from~\cite{dreher2024learning} (``Fuzzy + Heuristic'').
Since our approach is a full search that maximizes individual scores while ensuring contradiction-free assignments, we assume its result given all demonstrations to be a valid ground truth.
This evaluation thus shows how quickly each approach converges to this desired assignment.
Using the \textmueslitask task of the \gls{bimacs}, the model receives one demonstration at a time, and after each the constraints inferred by all methods are compared to this ground truth in a binary classification.
We repeat this process in $100$ learning scenarios, each varying the order in which the model receives demonstrations, and report the mean $F_1$ score and standard deviation.
\begin{figure}[!t]
    \vspace*{-1mm}
    \centering
    \includegraphics[width=.9\linewidth]{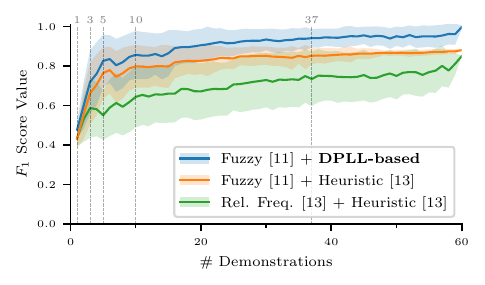}
    \vspace*{-3mm}
    \caption{%
        Mean $F_1$ scores $\pm$ one standard deviation for an increasing number of demonstrations across $100$ learning scenarios of two baselines compared to our approach (\gls{dpll}-based).}\label{fig:dpll-eva}
    \vspace*{-1mm}
\end{figure}
\begin{figure}[!t]
    \vspace*{-1mm}
    \centering
    \includegraphics[width=.9\linewidth]{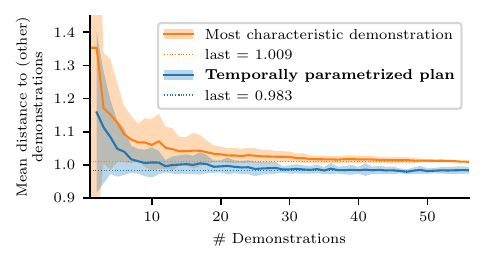}
    \vspace*{-3mm}
    \caption{%
        Mean distance $\pm$ one standard deviation of a temporally parametrized plan generated with our approach to all demonstrations. 
        The baseline is the distance of the most characteristic demonstration to all other demonstrations.}\label{fig:subtask-demonstration-distance-pour-banana}
    \vspace*{-1mm}
\end{figure}
\evalresults The results in Fig.~\ref{fig:dpll-eva} show that our \gls{dpll}-based approach consistently outperforms both methods, especially for a small number of demonstrations.
This can be attributed mainly to the fact that the \gls{dpll}-based approach does not rely on rule-based biases and seeks the most likely assignment given the demonstrations.
Already after ca. 37 demonstrations, almost all models of the \gls{dpll}-based approach outperform the heuristic baselines.
The mean inference time including temporal relationship assessment $\pm$ standard deviation of symbolic temporal task constraints over all demonstrations and learning scenarios is \SI{0.2 \pm 0.1}{\second} for Rel. Freq. + Heuristic, \SI{127.0 \pm 25.7}{\second} for Fuzzy + Heuristic, and \SI{293.8 \pm 303.5}{\second} for our Fuzzy + \gls{dpll}-based approach.
Note that the inference times of our \gls{dpll}-based approach include the time needed to infer the fuzzy-logic-based scores of Allen relations, since these are required as input.

\subsection{Action Timing Quality Assessment}

\evalexperimentalsetup Quantifying how good a generated temporally parametrized plan is can be challenging, as there is no objective \emph{best} timing.
Thus, in this evaluation, we compare the generated temporally parametrized plan with all demonstrations.
Both our temporally parametrized plan and a demonstration are essentially the same data structure (cf. Section~\ref{ss:def-task-action-demonstration}).
Let $P_a \! = \! (\vartimedaction_{p_1}, \! \vartimedaction_{p_2}, \! \dots, \! \vartimedaction_{p_n})$ be the set of actions in our temporally parametrized plan and $D_a \! = \! (\vartimedaction_{d_1}, \! \vartimedaction_{d_2}, \! \dots, \! \vartimedaction_{d_n})$ the set of corresponding actions in the demonstration.
Then we can define the distance between $P_a$ and $D_a$ as the Euclidean norm of the vector of semantic action differences as a generalization of Equation~\ref{eq:timing-distance}:
\begin{equation}
    d(P_a, D_a) = 
        \min_\varscalar %
        \modeuclideannormm{ %
        \begin{pmatrix} \vartimedactionstart_{d_1} \\ \vartimedactionend_{d_1} \\ \vartimedactionstart_{d_2} \\ \vdots \\ \vartimedactionend_{d_n} \end{pmatrix} \! %
        - \! %
        \left( \!\! %
        \begin{pmatrix} \vartimedactionstart_{p_1} \\ \vartimedactionend_{p_1} \\ \vartimedactionstart_{p_2} \\ \vdots \\ \vartimedactionend_{p_n} \end{pmatrix} \! %
        + \! %
        \varscalar \! \begin{pmatrix} 1 \\ 1 \\ 1 \\ \vdots \\ 1 \end{pmatrix} \!\! %
        \right) %
        }%
    \!
    .
\end{equation}
The model receives one demonstration at a time, after which the whole approach presented in Fig.~\ref{fig:graphical-abstract} is executed again to obtain an updated parametrized plan with the exception of the task assignment being chosen manually.
This isolates the timing parametrization from the symbolic search in the sense of an ablation to ensure comparable results.
We repeat this process in $50$ learning scenarios and report the mean distance $d(P_a, D_a)$ and variance that the respective temporally parametrized plan $P_a$ has to all demonstrations compared to a baseline of the most characteristic demonstration $D_a$.
The most characteristic demonstration is defined as the demonstration with the smallest distance to all other demonstrations known to the model at that time.
A direct comparison to the approach based on univariate \glspl{gmm}~\cite{dreher2024learning} is not possible, since it cannot produce plans.
We use the \textmueslitask task of the \gls{bimacs}.
Since the demonstrations are complex and feature many different task modes (cf.~\cite[Table V]{dreher2022learning}), we compare the quality of the parametrization for the \texttask{add banana} subtask.
\evalresults Fig.~\ref{fig:subtask-demonstration-distance-pour-banana} shows that our approach reliably yields a smaller distance to each individual demonstration compared to the baseline, especially for a small number of demonstrations.
This means that our approach can derive a parametrization for the plan closer to all demonstrations than the most characteristic demonstration.
Additionally, our formulation also allows to further condition the \gls{gmm} to specific temporal parameters or parameter ranges, like limiting the duration of an action to a given interval.

\subsection{Full Task Plan Assessment}

\evalexperimentalsetup In this evaluation, we show the result after running our full approach as depicted in Fig.~\ref{fig:graphical-abstract} using two example tasks of the \gls{bimacs}, namely the \textmueslitask task and the \textcomponenttask task.
All demonstrations are provided at once.
\evalresults Fig.~\ref{fig:fully-parametrized-tasks} shows the resulting temporally parametrized plans generated by our approach.
For the \textmueslitask, $12$ task modes (subtask sequences) were identified in total, while \textcomponenttask exhibited only one; within individual subtasks, between $25$ and $\num{20117}$ modes of slight variations were found, most of which are unlikely.
Gaps of \SI{1}{\second} have been added between subtasks to demonstrate that from the inferred task constraints, we know that there is no tight temporal relationship that must be met, leaving the robot time to approach objects for grasping or similar.

\vspace*{-0.7mm}
\subsection{Showcase: Orchestrated Execution}

In this showcase, we demonstrate how entire executions of tasks are orchestrated using temporally parametrized plans both in simulation and on real robots.
For this, demonstrations of tasks from the \gls{bmds} \cite{krebs2021kit} and recordings of an angle grinder disassembly were used to derive
\begin{enumerate*}[label=(\roman*)]
    \item plans as shown in Fig.~\ref{fig:graphical-abstract} from the segmentation data, and
    \item a library of full-pose \glspl{vmp}~\cite{daab2024incremental} associated to each action from motion capture and kinesthetic recordings, respectively.
\end{enumerate*}
We arrange and parameterize the \glspl{vmp} as dictated by the plan, and play back the generated executions in simulation using trajectories from an IK solver considering human-likeness criteria \cite{meixner2024unifying} or on a real robot.
The attached video (\href{https://youtu.be/B9oH8GLKSYg}{\texttt{youtu.be/B9oH8GLKSYg}}) shows several successful executions of synchronized bimanual manipulation tasks using a library of existing unimanual \glspl{mp} to give a qualitative impression.

\vspace*{-0.7mm}
\section{Conclusion and Future Work}

We presented a unified approach to simultaneously learn temporal task structure and action timing.
We contribute a novel representation of timings between two actions in a 3-D space, provide new methods to find assignments of Allen relations to each pair of actions of a task, and propose a novel planning system that can generate temporally parametrized plans for execution considering symbolic and subsymbolic temporal task constraints.
A limitation is that it relies on the heuristic task decomposition method from our previous work~\cite{dreher2022learning} to identify subtasks prior to symbolic assignment.
Since the runtime of our \gls{dpll}-based search grows exponentially with the number of action pairs considered, an imperfect decomposition that yields overly large subtasks can cause runtime spikes.
In future work, we will investigate task decomposition methods to address this limitation.
In addition, we believe that a combination of emerging and assigned synchronization approaches is necessary for a dynamic and goal-oriented orchestration of bimanual actions.
Ultimately, more research on general task models that incorporate many task constraint modalities is needed.

\vspace*{-0.6mm}

\balance
\bibliographystyle{IEEEtran}
\bibliography{2026_multivariate_gmms_planning}

\end{document}

%% file: h2t_def.tex
\usepackage[per-mode=fraction, separate-uncertainty=true, multi-part-units=single]{siunitx}
\usepackage{comment}
\usepackage[normalem]{ulem}
\usepackage{xspace}
\usepackage{xcolor}
\usepackage{lastpage}
\usepackage{soul}
\let\labelindent\relax
\usepackage[inline]{enumitem}

\DeclareSIUnit\px{px}
\DeclareSIUnit\fps{fps}

\definecolor{OliveGreen}{RGB}{0,200,25}
\newcommand{\red}[1]{\textcolor{red}{#1}}

\newcommand{\darkgreen}[1]{\textcolor{OliveGreen}{#1}}

\newcommand{\ie}{i.\,e.,\xspace}
\newcommand{\eg}{e.\,g.,\xspace}

\newcommand{\etal}{et\,al.\xspace}

\newif\iffinal

\newcommand{\enablefinalversion}{\finaltrue}

\newcommand{\replaced}[2]{%
	\iffinal%
	#2%
	\else%
	\red{\ifmmode\text{\sout{\ensuremath{#1}}}\else\sout{#1}\fi}\darkgreen{#2}%
	\fi%
}
\newcommand{\removed}[1]{%
	\iffinal%
	\else%
	\red{\ifmmode\text{\sout{\ensuremath{#1}}}\else\sout{#1}\fi}%
	\fi%
}

%% file: figures/gmms_in_timing_space.pdf_tex
\begingroup%
  \makeatletter%
  \providecommand\color[2][]{%
    \errmessage{(Inkscape) Color is used for the text in Inkscape, but the package 'color.sty' is not loaded}%
    \renewcommand\color[2][]{}%
  }%
  \providecommand\transparent[1]{%
    \errmessage{(Inkscape) Transparency is used (non-zero) for the text in Inkscape, but the package 'transparent.sty' is not loaded}%
    \renewcommand\transparent[1]{}%
  }%
  \providecommand\rotatebox[2]{#2}%
  \newcommand*\fsize{\dimexpr\f@size pt\relax}%
  \newcommand*\lineheight[1]{\fontsize{\fsize}{#1\fsize}\selectfont}%
  \ifx\svgwidth\undefined%
    \setlength{\unitlength}{245.99053691bp}%
    \ifx\svgscale\undefined%
      \relax%
    \else%
      \setlength{\unitlength}{\unitlength * \real{\svgscale}}%
    \fi%
  \else%
    \setlength{\unitlength}{\svgwidth}%
  \fi%
  \global\let\svgwidth\undefined%
  \global\let\svgscale\undefined%
  \makeatother%
  \begin{picture}(1,0.72616182)%
    \lineheight{1}%
    \setlength\tabcolsep{0pt}%
    \put(0,0){\includegraphics[width=\unitlength,page=1]{gmms_in_timing_space.pdf}}%
    \put(0.26955202,0.65229938){\makebox(0,0)[rt]{\lineheight{1.14999998}\smash{\begin{tabular}[t]{r}$\vartimingspaceoffset$\end{tabular}}}}%
    \put(0.01791699,0.25012287){\makebox(0,0)[lt]{\lineheight{1.14999998}\smash{\begin{tabular}[t]{l}$\vartimingspacerepralength$\end{tabular}}}}%
    \put(0.88827513,0.33840898){\makebox(0,0)[lt]{\lineheight{1.14999998}\smash{\begin{tabular}[t]{l}$\vartimingspacereprblength$\end{tabular}}}}%
    \put(0.5219716,0.05578637){\makebox(0,0)[rt]{\lineheight{1.14999998}\smash{\begin{tabular}[t]{r}\scriptsize \gls{gmm} mode\end{tabular}}}}%
    \put(0.41851196,0.177187){\makebox(0,0)[rt]{\lineheight{1.14999998}\smash{\begin{tabular}[t]{r}\scriptsize observed timing\end{tabular}}}}%
  \end{picture}%
\endgroup%

%% file: figures/allen_relations_in_timing_space.pdf_tex
\begingroup%
  \makeatletter%
  \providecommand\color[2][]{%
    \errmessage{(Inkscape) Color is used for the text in Inkscape, but the package 'color.sty' is not loaded}%
    \renewcommand\color[2][]{}%
  }%
  \providecommand\transparent[1]{%
    \errmessage{(Inkscape) Transparency is used (non-zero) for the text in Inkscape, but the package 'transparent.sty' is not loaded}%
    \renewcommand\transparent[1]{}%
  }%
  \providecommand\rotatebox[2]{#2}%
  \newcommand*\fsize{\dimexpr\f@size pt\relax}%
  \newcommand*\lineheight[1]{\fontsize{\fsize}{#1\fsize}\selectfont}%
  \ifx\svgwidth\undefined%
    \setlength{\unitlength}{245.97215427bp}%
    \ifx\svgscale\undefined%
      \relax%
    \else%
      \setlength{\unitlength}{\unitlength * \real{\svgscale}}%
    \fi%
  \else%
    \setlength{\unitlength}{\svgwidth}%
  \fi%
  \global\let\svgwidth\undefined%
  \global\let\svgscale\undefined%
  \makeatother%
  \begin{picture}(1,0.6367632)%
    \lineheight{1}%
    \setlength\tabcolsep{0pt}%
    \put(0,0){\includegraphics[width=\unitlength,page=1]{allen_relations_in_timing_space.pdf}}%
    \put(0.51487812,0.61782905){\makebox(0,0)[rt]{\lineheight{1.14999998}\smash{\begin{tabular}[t]{r}$\vartimingspaceoffset$\end{tabular}}}}%
    \put(0.32968121,0.3206962){\makebox(0,0)[rt]{\lineheight{1.14999998}\smash{\begin{tabular}[t]{r}$\vartimingspacerepralength$\end{tabular}}}}%
    \put(0.79636989,0.36749731){\makebox(0,0)[lt]{\lineheight{1.14999998}\smash{\begin{tabular}[t]{l}$\vartimingspacereprblength$\end{tabular}}}}%
    \put(0.87041063,0.5585336){\makebox(0,0)[lt]{\lineheight{1.14999998}\smash{\begin{tabular}[t]{l}\scriptsize\textallenbefore\end{tabular}}}}%
    \put(0.87041062,0.42024255){\makebox(0,0)[lt]{\lineheight{1.14999998}\smash{\begin{tabular}[t]{l}\scriptsize\textallenstarts\end{tabular}}}}%
    \put(0.87041063,0.28195147){\makebox(0,0)[lt]{\lineheight{1.14999998}\smash{\begin{tabular}[t]{l}\scriptsize\textallenduring\end{tabular}}}}%
    \put(0.87041063,0.21280595){\makebox(0,0)[lt]{\lineheight{1.14999998}\smash{\begin{tabular}[t]{l}\scriptsize\textallenequals\end{tabular}}}}%
    \put(0.25762658,0.21280595){\makebox(0,0)[rt]{\lineheight{1.14999998}\smash{\begin{tabular}[t]{r}\scriptsize\textallenoverlappedby\end{tabular}}}}%
    \put(0.25762661,0.42024255){\makebox(0,0)[rt]{\lineheight{1.14999998}\smash{\begin{tabular}[t]{r}\scriptsize\textallenfinishedby\end{tabular}}}}%
    \put(0.25762659,0.48938808){\makebox(0,0)[rt]{\lineheight{1.14999998}\smash{\begin{tabular}[t]{r}\scriptsize\textallenmeets\end{tabular}}}}%
  \end{picture}%
\endgroup%

%% file: figures/gmm_conditioning.pdf_tex
\begingroup%
  \makeatletter%
  \providecommand\color[2][]{%
    \errmessage{(Inkscape) Color is used for the text in Inkscape, but the package 'color.sty' is not loaded}%
    \renewcommand\color[2][]{}%
  }%
  \providecommand\transparent[1]{%
    \errmessage{(Inkscape) Transparency is used (non-zero) for the text in Inkscape, but the package 'transparent.sty' is not loaded}%
    \renewcommand\transparent[1]{}%
  }%
  \providecommand\rotatebox[2]{#2}%
  \newcommand*\fsize{\dimexpr\f@size pt\relax}%
  \newcommand*\lineheight[1]{\fontsize{\fsize}{#1\fsize}\selectfont}%
  \ifx\svgwidth\undefined%
    \setlength{\unitlength}{245.97212724bp}%
    \ifx\svgscale\undefined%
      \relax%
    \else%
      \setlength{\unitlength}{\unitlength * \real{\svgscale}}%
    \fi%
  \else%
    \setlength{\unitlength}{\svgwidth}%
  \fi%
  \global\let\svgwidth\undefined%
  \global\let\svgscale\undefined%
  \makeatother%
  \begin{picture}(1,0.63676322)%
    \lineheight{1}%
    \setlength\tabcolsep{0pt}%
    \put(0,0){\includegraphics[width=\unitlength,page=1]{gmm_conditioning.pdf}}%
    \put(0.66439412,0.56235328){\makebox(0,0)[lt]{\lineheight{1.14999998}\smash{\begin{tabular}[t]{l}\scriptsize \gls{gmm} mode\end{tabular}}}}%
    \put(0.72023393,0.4604613){\makebox(0,0)[lt]{\smash{\begin{tabular}[t]{l}\scriptsize most likely timing \\[-2pt] \scriptsize given \textallenfinishedby\end{tabular}}}}%
    \put(0.97843452,0.26054214){\makebox(0,0)[rt]{\lineheight{1.14999998}\smash{\begin{tabular}[t]{r}$\vartimingspacereprblength$\end{tabular}}}}%
    \put(0.01970987,0.18571867){\makebox(0,0)[lt]{\lineheight{1.14999998}\smash{\begin{tabular}[t]{l}$\vartimingspacerepralength$\end{tabular}}}}%
    \put(0.60168395,0.61132931){\makebox(0,0)[rt]{\lineheight{1.14999998}\smash{\begin{tabular}[t]{r}$\vartimingspaceoffset$\end{tabular}}}}%
    \put(0.74387336,0.1168143){\makebox(0,0)[lt]{\lineheight{1.14999998}\smash{\begin{tabular}[t]{l}\scriptsize\textallenfinishedby area\end{tabular}}}}%
  \end{picture}%
\endgroup%